\documentclass[runningheads]{cl2emult}

\usepackage{makeidx}  
\usepackage{graphicx} 
\usepackage{subeqnar} 
\usepackage{multicol} 
\usepackage{cropmark} 
\usepackage{math}     
\makeindex            

\begin{document}
\title*{A Computational Study of Genetic Crossover Operators for Multi-Objective
Vehicle Routing Problem with Soft Time Windows}
\toctitle{A Computational Study of Genetic Crossover Operators for
Multi-Objective Vehicle Routing Problem with Soft Time
Windows}
%
%
\titlerunning{Genetic Crossover Operators}
%
\author{Martin Geiger}
\authorrunning{M. Geiger}
%
%
\institute{University of Hohenheim, Production and Operations
Management Department, 70593 Stuttgart, Germany}

\maketitle              

\begin{abstract}
The article describes an investigation of the effectiveness of
genetic
        algorithms for multi-objective combinatorial optimization (MOCO) by
        presenting an application for the vehicle routing problem with soft
        time windows. The work is motivated by the question, if and how the
        problem structure influences the effectiveness of different
        configurations of the genetic algorithm. Computational results are
        presented for different classes of vehicle routing problems, varying
        in their coverage with time windows, time window size, distribution and
        number of customers. The results are compared with a simple, but
        effective local search approach for multi-objective combinatorial
        optimization problems.
\end{abstract}

\section{Introduction}
The requirements of modern logistics are widespread. It seems
natural that for practical problems several objectives have to be
optimized. The quality of service and products and the
corresponding cost are only a small subset of
necessary objectives that should be considered \cite{partyka2000article}.\\
The distribution of products, a part of the supply chain in
logistics, can be modelled as a vehicle routing problem (VRP). The
objective is to find a cost-minimizing set of routes from a depot
to serve a number of customers with known demands. Each customer
is serviced by exactly one of the vehicles which
capacities are not allowed to be exceeded.\\
The model of the vehicle routing problem with time windows (VRPTW)
generalizes the basic problem by introducing time windows for each
customer, defining an earliest and a latest possible time of
service. In the case of the vehicle routing problem with soft time
windows (VRPSTW), violation of these time windows is possible and
does not lead to infeasibility of the solution
\cite{balakrishnan1993article}. In our approach, the possible
violations of the time windows are modelled by introducing
objective functions, leading to a multi-objective model that is
able to describe the requirements of practical problems more
detailed compared to single-criterion optimization approaches. The
goal is to find all solutions minimizing the defined objective
functions. As these functions are often conflicting, the concept
of Pareto optimality is used to determine appropriate solutions.\\
The VRPTW \cite{solomon1987article}, as well as the VRP
\cite{lenstra1981article}, has been shown to be NP-hard in terms
of complexity. In combination with the practical relevance, this
is one of the reasons why there is still a lot of ongoing
research. Earlier approaches include heuristics and optimization
algorithms. Heuristics are focused on tour construction
\cite{solomon1987article}, tour improvement
\cite{savelsbergh1985article} or both in parallel
\cite{potvin1993article,russell1977article}.\\
Exact optimization methods successfully apply branch-and-bound
methods \cite{kolen1987article} or integer linear programming
based on Lagrangian relaxations
\cite{desrochers1992article,fisher1994article,fisher1997article}.
However, only smaller problem instances up to 50 customers can be
solved
reliable \cite{kohl1997techreport}.\\
More recent, metaheuristic strategies have been applied to the VRP
and the VRPTW. They include Simulated Annealing
\cite{chiang1996article,osman1993article}, Tabu Search
\cite{semet1993article,taillard1997article}, Genetic Algorithms
(GAs) \cite{thangiah1995incollection,thangiah1998incollection},
Ant Colony Optimization \cite{bullnheimer1999incollection}, hybrid
approaches \cite{thangiah1994techreport,thangiah1995techreport}
and newer local search
concepts like e.g. Guided Local Search \cite{kilby1999incollection}.\\
Nevertheless, the focus of the treated models is in the observed
cases on minimizing a single objective, the cost of the solution,
by minimizing the distances travelled by the used vehicles, they
do not address a more practical, multi-objective formulation with
a relaxation of the restrictive time windows. On the other side,
there is an increasing interest on applying evolutionary based
optimization techniques to multi-objective problems. Evolutionary
algorithms are regarded to be well suited for multi-objective
problems since a set of alternatives is used in the optimization
process and the goal is the approximation of a set of Pareto
optimal solutions. The idea is the convergence of the whole
population
towards the efficient frontier.\\
Since a first sketch of an idea by Goldberg
\cite{goldberg1989book}, most of the present work on
multi-objective optimization using evolutionary algorithms are
focused on the integration of the objective functions in the
calculation of the fitness values of the solutions, ranging from
scalarizing functions to Pareto-based techniques. Here to mention
are VEGA of Schaffer \cite{schaffer1985inproceedings2}, MOGA of
Fonseca and Fleming \cite{fonseca1993inproceedings}, NPGA of Horn
et al. \cite{horn1994inproceedings}, NSGA of Srinivas and Deb
\cite{srinivas1994article} and SPEA of Zitzler
\cite{zitzler1999phdthesis}. These approaches include the ideas of
fitness functions/fitness sharing, niching, mating restrictions
and elitism to define algorithms that maintain diversity within
the population, overcome local optima and finally converge towards
the set of efficient alternatives. For a detailed overview
concerning these topics, the interested reader is referred to
Coello and Van Veldhuizen
\cite{coello1999article,vanveldhuizen2000article}. Although big
progresses were made and a huge variety of applications were
already successfully developed \cite{coello2001website}, an
application to the vehicle routing problem under multiple
objectives is still missing. To define an appropriate algorithm,
the interchange between the problem structure, the configuration
and the behavior of the algorithm has to be studied. Especially
for the correct use of crossover operators in multi-objective
optimization problems, computational results are only for a few
examples available
\cite{brizuela2001inproceedings}.\\
This paper fills the gap in the described field of research. A
practical, multi-objective model of a VRPSTW is presented and a
genetic algorithm is defined to solve different problem instances,
varying in their time window coverage, time window size, number
and distribution of customers. Several crossover operators are
tested and computational results are given for the different
instances. To compare the results obtained by the GA, a simple
local search approach for multi-objective combinatorial
optimization is
presented.\\

\section{The multi-objective vehicle routing problem with soft time windows}
\subsection{Multi-objective optimization}\label{section_moo}
The goal of a multi-objective optimization is to
\begin{equation}\label{MOP}
    ``\min" G(R)=\big{(}g_{1}(R),...,g_{k}(R)\big{)}
\end{equation}
$R\in\Omega$ is a solution of the problem and belongs to the set
of all feasible solutions $\Omega$. The objective functions
$g_{1}(R),...g_{k}(R)$ map the decision $R$ into the objective
space, leading to a objective vector $G(R)$. As often conflicting
objective functions $g_{k}(R)$ are considered, minimization does
not lead to a single optimal solution but is understood in the
sense of efficiency (or Pareto optimality)
\cite{vanveldhuizen2000article}.

\begin{definition}\label{def1}
    An objective vector $G(R)$ is said to dominate $G(R')$,
    if $g_{i}(R)\leq g_{i}(R')\forall i\in \{ 1,...,k\}\wedge \exists i\in \{1,...,k\}|g_{i}(R)<g_{i}(R')$.
    We denote the domination of a vector $G(R)$ to the vector $G(R')$ with $G(R)\prec G(R')$.
\end{definition}
\begin{definition}\label{def2}
    A solution $R\in \Omega$ is said to be efficient or Pareto optimal,
    if $\neg\exists R'\in \Omega |R'\prec R$. Set set of all solutions
    fulfilling this property is called the Pareto set $P$.
\end{definition}

From the description of the multi-objective optimization problem
in equation \ref{MOP} we derive in combination with the
definitions \ref{def1} and \ref{def2} the goal to find all $R\in
P$.

\subsection{Problem description}
The vehicle routing problem with soft time windows can be
described as
follows:\\
A known number of customers have to be delivered from a depot with
a known amount of goods for which an unlimited number of
homogeneous vehicles is available. It is assumed that each
customer is visited by exactly one vehicle and a loading and a
travelling constraint exists for the vehicles. A soft time window
is associated with each customer, defining a desired earliest and
a latest time of service. Violation of these time windows does not
lead to infeasibility of the solution. With respect to the soft
nature of the time windows, it is assumed that service is done
immediately after the arrival of
the vehicle.\\
The objective of the problem is to maximize quality of service and
to minimize cost, such that the requirements of the customers and
the side-constraints are met. It is obvious, that the violation of
the time windows has to be minimized in order to achieve a high
quality of service. This can be done by minimizing the number of
time window violations and the time window violations itself,
measured in time units.\\
The cost consist of a fixed part, induced by the number of used
vehicles and
a variable part, caused by the route length and the travel time.\\

\subsection{Notation of the model}\label{section_notation}
The customers and the depot are modelled as vertices
$V=\{v_0,v_1,...,v_N\}$ in a graph $G=(V,A)$, connected by a set
of arcs $A$. In $V$, the depot is
indexed by 0 and the total number of customers is $N$.\\
With each arc $(v_i,v_j)$ from node $v_i$ to $v_j$, a nonnegative
travel time $t_{v_i,v_j}$, a distance $d_{v_i,v_j}$ and a cost
$c_{v_i,v_j}$ is associated. It is assumed that
$t_{v_i,v_j}=d_{v_i,v_j}=c_{v_i,v_j}\quad
\forall (v_i,v_j)\in A$.\\
$u_{v_i}$: unloading time at customer $v_i,v_i\in V$.\\
$d_{v_i}$: demand of customer $v_i,v_i\in V$.\\
$d^{max}$: capacity constraint of the vehicles.\\
$[a_{v_i}, b_{v_i}]$: time window at customer $v_i,v_i\in V$.
$a_{v_i}$ defines the desired earliest, $b_{v_i}$ the latest time
of service. It is
necessary that $b_{v_i}\geq a_{v_i}\quad \forall v_i\in V$.\\
$t_{v_i}^a$: arrival time of the vehicle at customer $v_i,v_i\in V$.\\
$r_k=\big{[}[1]_k,[2]_k,...,[n_k]_k\big{]}$: route $r_k$,
representing $n_k$ customers in a given sequence. To assure, that
each customer is included in exactly one route, it is necessary
that $r_k\cap r_{k'}=\emptyset \quad
\forall k\not= k'$ and $\mid \cup r_k\mid = \mid V\setminus \{v_0\}\mid$.\\
$R=\{r_1,...,r_m\}$: set of routes of the solution. The number of
routes $m$
can be greater than one and does not have to be equal in every solution.\\
The time $t(r_k)$ to travel route $r_k$ can now be obtained by
calculating
\begin{equation}
    t(r_k)=a_{v_0}+t_{v_0,[1]_k}+t_{[n_k]_k,v_0}+\sum\limits_{[i]_{k}=1}^{n_{k}-1}(t_{[i]_k,[i+1]_k}+u_{[i]_k})+u_{[n_k]_k}
\end{equation}
As the time window $[a_{v_0},b_{v_0}]$ defines the interval in
which the vehicles are available, it is necessary that
\begin{equation}\label{constraint_time}
    t(r_k)\leq b_{v_0}\quad\forall r_k\in R\\
\end{equation}
The quantity $q(r_k)$ of transported goods on route $r_k$ is given
by
\begin{equation}
    q(r_k)=\sum\limits_{i=1}^{n_k}d_{v_i}
\end{equation}
The capacity constraint must be met, so
\begin{equation}\label{constraint_quantity}
    q(r_k)\leq d^{max}\quad\forall r_k\in R
\end{equation}
$w(v_i)$ measures the time window violation of customer
$v_i,v_i\in V$.
\begin{equation}
    w(v_i)=\max\big{(}\max(0;a_{v_i}-t_{v_i}^a);\max(0;t_{v_i}^a-b_{v_i})\big{)}
\end{equation}
The fact, that the time window of customer $v_i$ is violated is
indicated by $u(v_i)$:
\begin{equation}
    u(i)=\left\{\begin{array}{r@{\quad :\quad}l}
        1 & w(i)>0\\
        0 & w(i)=0\\
    \end{array} \right.
\end{equation}
The objective functions are given by:
\begin{equation}\label{min_routes}
    g_1(R)=\sum\limits_{k=1}^{\mid R\mid}t(r_k)
\end{equation}
\begin{equation}\label{min_vehicles}
    g_2(R)=\mid R \mid
\end{equation}
\begin{equation}\label{min_twviol}
    g_3(R)=\sum\limits_{i=1}^{N}w(v_i)
\end{equation}
\begin{equation}\label{min_notwviol}
    g_4(R)=\sum\limits_{i=1}^{N}u(v_i)
\end{equation}
The objective functions (\ref{min_routes}) and
(\ref{min_vehicles}) minimize the costs associated with the
solution. Function (\ref{min_routes}) minimizes the length of the
routes, travelled by the vehicles. In addition to that, function
(\ref{min_vehicles}) minimizes the number of used vehicles. To
maximize the provided service it is desired to minimize occurring
time window violations. The objective functions (\ref{min_twviol})
and (\ref{min_notwviol}) express this circumstance by the measured
deviations in time units and by the number of violated time windows.\\
According to the definitions (\ref{def1}) and (\ref{def2}) given
in section \ref{section_moo}, the goal is to find all $R\in P$.

\newpage
\section{A genetic approach}
\subsection{Encoding technique, initialization}\label{section_encoding}
We use a string of $N$ genes to encode the solutions of the
problem. Each gene represents a customer of the problem and the
sequence of them corresponds to the sequence of visiting the
customers. A possible string, e.g. \ttfamily "5 2 1 3 7 9 8 4 6
10" \normalfont is then partitioned to a set of tours, e.g.
$r_{1}=[v_{5},v_{2},v_{1}],r_{2}=[v_{3},v_{7},v_{9},v_{8}],r_{3}=[v_{4},v_{6},v_{10}]$,
assuring feasibility with respect to the defined side constraints
(\ref{constraint_time}) and (\ref{constraint_quantity}) in section
\ref{section_notation}. The decoding technique inserts the next
stored customer of the representation in the current tour as long
as the side
constraints are not being violated.\\
The initial population inherits $n^{pop}$ individuals $i$,
consisting of randomly generated strings. To assure a sufficient
potential of genetic information and avoid premature convergence,
$n^{pop}$ is chosen equal to 500.

\subsection{Fitness function}
Most investigations on genetic algorithms for multi-objective
optimization are based on examining the effectiveness of certain
approaches of fitness functions. As we have a different focus, we
restrict ourselves to the use of a single function, similar to the
the well known fitness function
proposed by Fonseca and Fleming \cite{fonseca1993inproceedings}.\\
For individual $i$ of the current population, the number of
individuals whose corresponding objective vectors dominate the
objective vector of $i$ is indicated by $\xi_{i}$. All currently
not being dominated alternatives receive $\xi_{i}=0$. The fitness
$f(i)$ of individual $i$ is derived using a linear transformation
with $\xi^{max}=\max\limits_{i}(\xi_{i})$ and two external given
parameters $f^{min}$ and $f^{max}$:
\begin{equation}
    f(i)=f^{max}-\frac{\xi_{i}(f^{max}-f^{min})}{\xi^{max}}
\end{equation}
The parameters $f^{min}$ and $f^{max}$ are set to $f^{min}=1$ and
$f^{max}=100$.

\subsection{Genetic recombination operators}
As we want to study the effectiveness of different crossover
operators, several of them are implemented and tested. From the
first approaches of genetic crossover operators, improvements were
made by incorporating problem specific knowledge into the
recombination process. This is in the case of the travelling
salesman problem as well as in the case of the vehicle routing
problem e.g. done by considering the distances of the customers to
each other and selecting between two possible genes the route
minimizing
part.\\
As we deal with more than one objective, problem specific
orientation during the crossover process is not as obvious. We
therefore restrict the investigation in this first approach on
problem independent working
operators.\\
Among the best studied crossover operators for permutation based
encodings are the ``partially mapped crossover" (PMX)
\cite{goldberg1985inproceedings2}, the ``order based crossover"
(OBX) \cite{syswerda1991incollection} and the ``uniform order
based crossover" (UOBX) \cite{davis1991incollection}. We
apply them with a crossover probability of $p^{cross}=1$.\\
Crossover operators introduce implicit mutations to the offspring.
To distinguish between the true results obtained by crossover, no
mutation operator is used in the basic test runs. However, for
comparison reason a simple ``swap mutation" exchanging two
randomly picked genes is examined in a separate test run in
combination with the UOBX as this crossover operator is being
described as the best among the problem independent operators for
the considered problem type. In this configuration
(UOBX$\wedge$2EX) , it is applied with a probability of
$p^{mut}=0.1$ for the whole individual. Thus, the probability of
an exchange of a single gene is $\frac{2p^{mut}}{N}$.

\subsection{Elitism}
It has been shown, that elitism plays in general an important role
in genetic algorithms and in particular in multi-objective GAs,
see e.g. \cite{zitzler1999phdthesis}. Various approaches exist to
implement an elitist strategy, ranging from preserving the $k$
best individuals in
the population to storing the best individuals in an external set.\\
We use a simple, but effective strategy to obtain an elitist
genetic algorithm by using overlapping populations. This principle
has been introduced as the so called ``steady state" genetic
algorithm. From each generation to the next, the population is
preserved and new individuals are only inserted if they improve
the average quality of the population. Duplicates are not allowed,
including duplicates of encoded strings as well as duplicates of
the corresponding solutions in the decision space. The improvement
of the population is in our case measured by the $\xi_{i}$-values.
This means that the complexity of the necessary calculations to
obtain the $\xi_{i}$-values by pairwise comparison is $2n^{pop}$.

\subsection{Termination criterion}
The algorithm can terminate after a defined time, a given number
of calculations or after the algorithm has converged to a given
goal and does not produce any improvements for a defined time or
number of calculations. It is certainly interesting to perform
calculations until a convergence to a local or global optimal
front is reached. The termination criterion is chosen according to
this goal. The genetic algorithm stops after 10,000 iterations
with no found new individual $i$ having a $\xi_{i}=0$.

\section{A local search approach}
A simple local search approach is used to compare the results
obtained by the genetic algorithm. It is based on the encoding
scheme of section
\ref{section_encoding} and can be described as follows:\\
\ttfamily
\begin{tabbing}
AA\= \kill
1 initialize\\
 \> 1.1 generate starting solution $R$\\
 \> 1.2 encode starting solution $R$\\
 \> 1.3 set $P^{approx}=\{R\}$\\
2 repeat until neighborhood of $R\in P^{approx}$ is investigated \\
 \> $\forall R\in P^{approx}$\\
 \> 2.1 select new solution $R\in P^{approx}$\\
 \> 2.2 generate neighborhood of $R$\\
 \> 2.3 update $P^{approx}$\\
 \> 2.4 set neighborhood of $R$ as investigated\\
\end{tabbing}
\normalfont The algorithm starts from a random generated solution.
The calculation of the initial solution is done as in the
initialization procedure of the genetic algorithm to assure proper
comparison results. A set $P^{approx}$ is used to store the set of
currently not dominated solutions. According to the given
definitions (\ref{def1}) and (\ref{def2}) in section
\ref{section_moo} it has to be noticed that these solutions do not
necessarily fulfill the requirements of Pareto optimality.
However, as we do not know the real efficient solutions and the
true Pareto set $P^{true}$, we consider the nondominated solutions
as an approximation
of the Pareto set $P^{true}$.\\
The algorithm then starts in step \ttfamily 2 \normalfont to
investigate the neighborhood of the solutions in $P^{approx}$. A
solution $R$ is randomly selected from the set $P^{approx}$ and
the neighboring solutions are calculated. Procedure \ttfamily 2.3
\normalfont then inserts new nondominated solutions of the
neighborhood of $R$ into the set $P^{approx}$
and removes dominated ones.\\
The local search procedure terminates when the neighborhood all
solutions $R\in P^{approx}$ has been investigated and as result a
set of locally
optimal solutions with respect to the used neighborhood is obtained.\\
An approach based on the neighborhood proposed by Lin is used to
perform local search. Every possible substring of the encoding is
extracted, reversed and reinserted, resulting in
$\frac{N(N-1)}{2}$ calculations. The basic ideas of this method
come from the local search descent method, incorporating
multi-objective storage of nondominated solutions. We call this
approach therefore multiple-objective local search descent
(MOLSD).

\section{Computational experiments}
\subsection{Test problems}
To investigate the influence of the problem structure on the
effectiveness of different configurations of the algorithms
proposed above, a set of test
instances incorporating different characteristics of the problem is necessary.\\
For that purpose, 40 test instances based on the data given from
Solomon \cite{solomon1987article} were created. Each instance can
be described using an $\alpha;\beta;\gamma;\delta$-classification
scheme, representing the following configurations:
\begin{itemize}
  \item $\alpha$: Distribution of the customers.\\We consider test instances
  with clustered data sets ($\alpha=C$) and with random distributed
  customers ($\alpha=R$).
  \item $\beta$: Number of customers.\\Problems with 20 and 30 customers were
  defined.
  \item $\gamma$: Coverage with time windows.\\This attribute measures the
  relation of the number of customers having time windows to the number of customer without time windows.
  \item $\delta$: Average time window size.\\The size of the time window at
  customer $v_{i}$ is defined as $b_{v_{i}}-a_{v_{i}}$.
\end{itemize}
Example: The string ``C;20;0.70;60" describes the instance with a
clustered distribution of 20 customers, having a density of 70\%
with time windows and
an average time window size of 60 time units.\\
We expect the problems with random distribution of the customers
and a high density of small time windows to be more complex to
solve than problems with less and bigger time windows. This should
affect the speed of convergence
of the algorithms and the quality of the obtained solutions.\\
The neighborhood of the defined problems with 30 customers
contains 2.29 times more alternatives than the one of the problems
with 20 customers. An increase of the necessary evaluations of at
least this factor is being expected.

\subsection{Measures of effectiveness}
Each of the performed test runs leads to an approximation
$P^{approx}$ of the set of efficient alternatives $P$. Due to the
complexity of the problem, we do not know the true Pareto set $P$.
However, to obtain a measure of effectiveness, we compute a
reference set $P^{ref}$ including all the found nondominated
solutions during the test runs. The effectiveness of the
approximations is then oriented on the approximation of this
reference
set $P^{ref}$.\\
To determine the distance between two sets of alternatives, a
distance function measuring the distance of the underlying
elements according to the the proposal of Czyzak and Jaszkiewicz
\cite{czyzak1998article} is used. The distance measure can be
described as a weighted achievement scalarizing function, using
$w_{j}=\frac{1}{\Delta_{j}}$ as a weight for the goals based on
the spread $\Delta_{j}$ of the objective function values of
objective $j$ in the reference set $P^{ref}$. For an alternative
$x\in P^{approx}$ to an alternative $y\in P^{ref}$, the distance
is measured by equation (\ref{c_x_y}).
\begin{equation}\label{c_x_y}
    c(x,y)=\max\limits_{j=1,...,k}\Big{\{}0;w_{j}\big{(}g_{j}(x)-g_{j}(y)\big{)}\Big{\}}
\end{equation}
Two approaches are used to obtain the distance of the
approximation $P^{approx}$ to the reference set $P^{ref}$. We
measure the average distance by equation (\ref{d1}) and the
worst-case distance by equation (\ref{d2}).
\begin{equation}\label{d1}
    d_{1}=\frac{1}{\mid P^{ref} \mid}\sum\limits_{y\in P^{ref}}\min\limits_{x\in M }\big{\{}c(x,y)\big{\}}
\end{equation}
\begin{equation}\label{d2}
    d_{2}=\max\limits_{y\in P^{ref}}\big{\{}\min\limits_{x\in M}\{c(x,y\}\big{\}}
\end{equation}
As our goal is to minimize the distance between the obtained
approximations $P^{approx}$ and the reference set $P^{ref}$, the
distances $d_{1}$ and
$d_{2}$ are to be minimized.\\
To come to stable and reliable conclusions, average values of
$d_{1}$ and $d_{2}$ of several test runs with the same
configuration are computed.

\subsection{Results}
The test runs were performed on a personal computer, equipped with
an 1 GHz Intel Pentium III Processor and 384 MB RAM. Each
algorithmic configuration was executed 100 times for each test
problem in order to obtain stable results. The average number of
evaluations of the test runs and the performance of the
implemented heuristics is given in table
\ref{tbl_number_evaluations} in the appendix. Resulting from the
lower computational effort, the MOLSD is able to evaluate about
4.70 times more solutions per second than the GAs. Each second,
1,385 evaluations were computed
by the MOLSD.\\
As expected, all algorithms tend to converge slower in instances
with dense and small time windows and random distributed
customers. It is worth to mention, that the increase of the
necessary evaluations of the MOLSD from a $\beta=20$ to a
$\beta=30$ was about factor 9.14, a huge increase compared to the
additional computations of the GAs, having factors of 1.80 for the
PMX,
1.96 for the OBX, 1.78 for the UOBX and 2.63 for the UOBX$\wedge$2EX.\\
The results, given in table \ref{tbl_d1} and \ref{tbl_d2} in the
appendix indicate a strong dependency of the approximation quality
on the distribution of the customers $\alpha$. The examined
crossover operators PMX, OBX, UOBX itself as well as the genetic
algorithm including the swap mutation UOBX$\wedge$2EX were in most
instances with an $\alpha=C$ not able to outperform the simple
local search approach MOLSD. Both defined measures $d_{1}$ and
$d_{2}$ are affected by this property and the distance between the
values of the MOLSD to the values of the GAs is increasing with
falling $\gamma$ or rising $\delta$. We conclude, that VRPSTW with
a structured distribution of the customers are more easy to be
solved with simple local search while the examined configurations
of the genetic algorithm tend to get stucked in local optima. This
effect grows with a declining complexity of the underlying
problem, expressed by a low coverage with time windows $\gamma$ or
wide time windows $\delta$. Similar results have already been
reported for the single objective VRPTW
\cite{blanton1993inproceedings} and
can be confirmed for the multi-objective formulation.\\
For VRPSTW with a random distribution of the customer we observe
the contrary. In most test instances, the best results for the
average distance $d_{1}$ are obtained by the UOBX$\wedge$2EX.
However, the advantage of the UOBX$\wedge$2EX to the MOLSD is
decreasing with falling $\gamma$ or increasing $\delta$. For some
problems with large time windows $\delta$ and a small coverage
$\gamma$,
e.g. the ``R;30;1.00;80/95/115", MOLSD produces better results.\\
The results of $d_{2}$ indicate, that the average worst-case
deviation of the approximation $P^{approx}$ to the reference set
$P^{ref}$ is superior in the approach of the MOLSD, except for
some smaller instances with $\beta=20$ and
the hard ``R;30;1.00;10".\\
A closer investigation of the results shows, that the superior
results of the UOBX$\wedge$2EX is due to the use of the swap
mutation operator. Comparing the results for the crossover-only
GAs, it can be seen that with respect to the distances $d_{1}$ and
$d_{2}$ OBX and UOBX perform in most instances better than PMX. On
the other hand, a clear distinction between OBX and UOBX can't be
concluded.\\
The results show a strong influence of the mutation operator on
the effectiveness for the genetic algorithm. Although the GAs lead
to weaker results in clustered data sets than the proposed local
search approach, this behavior seems to be typical for the
presented multi-objective optimization problem.

\section{Conclusions}
The goal of the paper was to investigate the influence of the
problem structure on the effectiveness of different configurations
of a genetic algorithm. A multi-objective formulation of the
vehicle routing problem with soft time windows has been presented,
describing the requirements of practical logistics more detailed
than single criterion models. For comparison reasons, an approach
of a local search approach incorporating storage of multiple
nondominated solutions was proposed.\\
The results show, that the problem structure, especially the
distribution of the customers $\alpha$, is crucial for the
behavior of the investigated algorithms. The genetic algorithms
outperform the local search approach especially well in complex
problems with dense and small time windows and a random
distribution of the customers. However, the influence of the
mutation operator seems to be stronger than expected and described
in other publications
on GAs. \\
An important role of simple (local search based) mutations can
indicate, that the studied crossover operators themselves are
comparable weak for the multi-objective formulation of the problem
as they do not recombine the desirable structures of the
underlying model. Nevertheless, specific formulations of
particular multi-objective operators are still missing. A
combination of genetic operators with local search heuristics is
consequently a logical conclusion of the obtained results.

\bibliography{../lit_bank,../lit_bank_ibl,../lit_bank_nv}

\begin{thebibliography}{10}

\bibitem{balakrishnan1993article}
Nagraj Balakrishnan.
\newblock Simple heuristics for the vehicle routeing problem with soft time
  windows.
\newblock {\em Journal of the Operational Research Society}, 44(3):279--287,
  1993.

\bibitem{blanton1993inproceedings}
Joe~L. {Blanton Jr.} and Roger l.~Wainwright.
\newblock Multiple vehicle routing with time and capacity constraints using
  genetic algorithms.
\newblock In Forrest \cite{forrest1993proceedings}, pages 452--459.

\bibitem{brizuela2001inproceedings}
C.~Brizuela, N.~Sannomiya, and Y.~Zhao.
\newblock Multi-objective flow-shop: Preliminary results.
\newblock In Eckart Zitzler, Kalyanmoy Deb, Lothar Thiele, and Carlos A.~Coello
  Coello, editors, {\em Evolutionary Multi-Criterion Optimization: First
  International Conference, EMO 2001}, pages 443--457, 2001.

\bibitem{bullnheimer1999incollection}
Bernd Bullnheimer, Richard~F. Hartl, and Christine Strauss.
\newblock Applying the ant system to the vehicle routing problem.
\newblock In Voss et~al. \cite{voss1999book}, chapter~20, pages 285--296.

\bibitem{chiang1996article}
Wen-Chyuan Chiang and Robert~A. Russel.
\newblock Simulated annealing metaheuristics for the vehicle routing problem
  with time windows.
\newblock {\em Annals of Operations Research}, 63:3--27, 1996.

\bibitem{coello2001website}
Carlos A.~Coello Coello.
\newblock List of references on evolutionary multiobjective optimization.
  http://www.lania.mx/\~{}ccoello/{EMOO/EMOO}bib.html.

\bibitem{coello1999article}
Carlos A.~Coello Coello.
\newblock A comprehensive survey of evolutionary-based multiobjetive
  optimization techniques.
\newblock {\em Knowledge and Information Systems}, 3:269--308, 1999.

\bibitem{czyzak1998article}
Piotr Czyzak and Andrej Jaskiewicz.
\newblock Pareto simulated annealing - a metaheuristic technique for
  multiple-objective combinatorial optimization.
\newblock {\em Journal of Multi-Criteria Decision Analysis}, 7:34--47, 1998.

\bibitem{davis1991incollection}
Lawrence Davis.
\newblock A genetic algorithms tutorial.
\newblock In {\em Handbook of genetic algorithms\/} \cite{davis1991book},
  chapter~I, pages 1--101.

\bibitem{davis1991book}
Lawrence Davis, editor.
\newblock {\em Handbook of Genetic Algorithms}.
\newblock Van Nostrand Reinhold, New York, 1991.

\bibitem{desrochers1992article}
Martin Desrochers, Jacques Desrosiers, and Marius Solomon.
\newblock A new optimization algorithm for the vehicle routing problem with
  time windows.
\newblock {\em Operations Research}, 40(2):342--354, 1992.

\bibitem{fisher1994article}
M.L. Fisher.
\newblock Optimal solution of vehicle routing problems using {K}-trees.
\newblock {\em Operations Research}, 42(4):626--642, 1994.

\bibitem{fisher1997article}
M.L. Fisher and K.O. Jornsten.
\newblock Vehicle routing with time windows: Two optimization algorithms.
\newblock {\em Operations Research}, 45(3):488--492, 1997.

\bibitem{fonseca1993inproceedings}
Carlos~M. Fonseca and Peter~J. Fleming.
\newblock Genetic algorithms for multiobjective optimization: Formulation,
  discussion and generalization.
\newblock In Forrest \cite{forrest1993proceedings}, pages 416--423.

\bibitem{forrest1993proceedings}
Stephanie Forrest, editor.
\newblock {\em Proceedings of the Fifth International Conference on Genetic
  Algorithms}, San Mateo, CA, 1993. Morgan Kaufmann Publishers.

\bibitem{goldberg1985inproceedings2}
David~E. Goldberg and Robert Lingle, Jr.
\newblock Alleles, loci, and the traveling salesman problem.
\newblock In Grefenstette \cite{grefenstette1985proceedings}, pages 154--159.

\bibitem{goldberg1989book}
D.E. Goldberg.
\newblock {\em Genetic Algorithms in Search, Optimization, and Machine
  Learning}.
\newblock Addison Wesley, Reading, Massachusetts, 1989.

\bibitem{grefenstette1985proceedings}
John~J. Grefenstette, editor.
\newblock {\em Proceedings of the First International Conference on Genetic
  Algorithms and their Applications}, Hillsdale, NJ, 1985. Lawrence Erlbaum
  Associates.

\bibitem{horn1994inproceedings}
J.~Horn, N.~Nafpliotis, and D.E. Goldberg.
\newblock A niched pareto genetic algorithm for multiobjective optimization.
\newblock In Z.~Michalewicz, editor, {\em Proceedings of the First IEEE
  Conference on Evolutionary Computation}, pages 82--87, Piscataway, New
  Jersey, 1994. IEEE Press.

\bibitem{kilby1999incollection}
Philip Kilby, Patrick Prosser, and Paul Shaw.
\newblock Guided local search for the vehicle routing problem with time
  windows.
\newblock In Voss et~al. \cite{voss1999book}, chapter~32, pages 473--486.

\bibitem{kohl1997techreport}
N.~Kohl, J.~Desrosiers, O.B.G. Madsen, M.M. Solomon, and F.~Soumis.
\newblock k-path cuts for the vehicle routing problem with time windows.
\newblock Technical Report G-97-19, Les Cahiers di GERAD, 1997.

\bibitem{kolen1987article}
A.W.J. Kolen, A.H.G. {Rinnooy Kan}, and H.W.J.M. Trienekens.
\newblock Vehicle routing with time windows.
\newblock {\em Operations Research}, 35(2):266--273, 1987.

\bibitem{lenstra1981article}
J.~Lenstra and A.~Rinnooy Kan.
\newblock Complexity of vehicle routing and scheduling problems.
\newblock {\em Networks}, pages 221--228, 1981.

\bibitem{osman1993article}
Ibrahim~Hassan Osman.
\newblock Metastrategy simulated annealing and tabu search algorithms for the
  vehicle routing problem.
\newblock {\em Annals of Operations Research}, 41:421--451, 1993.

\bibitem{partyka2000article}
J.G. Partyka and R.W. Hall.
\newblock On the road to service.
\newblock {\em OR/MS Today}, pages 26--35, August 2000.

\bibitem{potvin1993article}
J.-Y. Potvin and J.M. Rousseau.
\newblock A parallel route building algorithm for the vehicle routing and
  scheduling problem with time windows.
\newblock {\em European Journal of Operational Research}, 66:331--340, 1993.

\bibitem{russell1977article}
R.~Russel.
\newblock An effective heuristic for the {$M$}-tour traveling salesman problem
  with some side conditions.
\newblock {\em Operations Research}, 25:517--524, 1977.

\bibitem{savelsbergh1985article}
M.~Savelsbergh.
\newblock Local search in routing problems with time windows.
\newblock {\em Annals of Operations Research}, 4:285--305, 1985.

\bibitem{schaffer1985inproceedings2}
J.~David Schaffer.
\newblock Multiple objective optimization with vector evaluated genetic
  algorithms.
\newblock In Grefenstette \cite{grefenstette1985proceedings}, pages 93--100.

\bibitem{semet1993article}
Fr\'{e}d\'{e}ric Semet and Eric Taillard.
\newblock Solving real-life vehicle routing problems efficiently using tabu
  search.
\newblock {\em Annals of Operations Research}, 41:469--488, 1993.

\bibitem{solomon1987article}
Marius~M. Solomon.
\newblock Algorithms for the vehicle routing and scheduling problems with time
  window constraints.
\newblock {\em Operations Research}, 35(2):254--265, March-April 1987.

\bibitem{srinivas1994article}
N.~Srinivas and K.~Deb.
\newblock Multiobjective optimization using nondominated sorting in genetic
  algorithms.
\newblock {\em Evolutionary Computation}, 2(3):221--248, 1994.

\bibitem{syswerda1991incollection}
Gilbert Syswerda.
\newblock Schedule optimization using genetic algorithms.
\newblock In Davis \cite{davis1991book}, chapter~21, pages 332--349.

\bibitem{taillard1997article}
E.D. Taillard, F.~Badeau, M.~Gendreau, F.~Guertin, and J.-Y. Potvin.
\newblock A tabu search heuristic for the vehicle routing problem with soft
  time windows.
\newblock {\em Transportation Science}, 31:170--186, 1997.

\bibitem{thangiah1995incollection}
Sam~R. Thangiah.
\newblock Vehicle routing with time windows using genetic algorithms.
\newblock In Lance Chambers, editor, {\em Application Handbook of Genetic
  Algorithms: New Frontiers}, volume~2, pages 253--277. CRC Press, 1995.

\bibitem{thangiah1998incollection}
S.R. Thangiah.
\newblock Genetic algorithms, tabu search and simulated annealing methods for
  vehicle routing problems with time windows.
\newblock In Lance Chambers, editor, {\em Practical Handbook of Genetic
  Algorithms: Complex Structures}. CRC Press, 1998.

\bibitem{thangiah1994techreport}
S.R. Thangiah, I.H. Osman, and T.~Sun.
\newblock Hybrid genetic algorithms, simulated annealing and tabu search
  methods for vehicle routing problems with time windows.
\newblock Technical report, Institute of Mathematics and Statistics, University
  of Kent, United Kingdom, 1994.

\bibitem{thangiah1995techreport}
S.R. Thangiah, I.H. Osman, and T.~Sun.
\newblock Metaheuristics for vehicle routing problems with time windows.
\newblock Technical report, Institute of Mathematics and Statistics, University
  of Kent, United Kingdom, 1995.

\bibitem{vanveldhuizen2000article}
David~A. {Van Veldhuizen} and Gary~B. Lamont.
\newblock Multiobjective evolutionary algorithms: Analyzing the
  state-of-the-art.
\newblock {\em Evolutionary Computation}, 8(2):125--147, 2000.

\bibitem{voss1999book}
Stefan Voss, Silvano Martello, Ibrahim~H. Osman, and Catherine Roucairol,
  editors.
\newblock {\em META-HEURISTICS: Advances and Trends in Local Search Paradigms
  for Optimization}.
\newblock Kluwer Academic Publishers, Boston, Dordrecht, London, 1999.

\bibitem{zitzler1999phdthesis}
Eckart Zitzler.
\newblock {\em Evolutionary Algorithms for Multiobjective Optimization: Methods
  and Applications}.
\newblock PhD thesis, Computer Engineering and Networks Laboratory, Swiss
  Federal Institute of Technology, Z\"urich, Switzerland, 1999.

\end{thebibliography}
\bibliographystyle{plain}
\newpage
\begin{appendix}
\section{Appendix}

\begin{table}\centering\caption{Average number of
evaluations.}\label{tbl_number_evaluations}
\begin{tabular}{lrrrrr} \hline
$\alpha;\beta;\gamma;\delta$ & MOLSD & PMX & OBX & UOBX & UOBX$\wedge$2EX\\
\hline
C;20;1.00;60 & 38,730 & 108,750 & 54,375 & 75,625 & 80,000\\
C;20;0.70;60 & 43,643 & 98,125 & 80,625 & 98,125 & 95,625\\
C;20;0.45;60 & 35,488 & 103,750 & 82,500 & 89,375 & 96,250\\
C;20;0.30;60 & 26,938 & 71,875 & 56,875 & 64,375 & 76,875\\
C;20;1.00;120 & 36,814 & 105,625 & 50,625 & 66,875 & 48,125\\
C;20;1.00;180 & 42,047 & 175,625 & 63,750 & 45,000 & 138,750\\
C;20;1.00;240 & 26,125 & 73,750 & 55,625 & 46,875 & 58,125\\
C;20;1.00;360 & 23,978 & 74,375 & 36,250 & 35,000 & 34,375\\
R;20;1.00;10 & 90,060 & 425,625 & 381,875 & 438,125 & 474,375\\
R;20;0.70;10 & 87,199 & 385,625 & 398,125 & 353,125 & 411,875\\
R;20;0.45;10 & 90,011 & 386,875 & 301,875 & 303,750 & 433,125\\
R;20;0.30;10 & 46,364 & 196,250 & 204,375 & 238,750 & 231,875\\
R;20;1.00;30 & 90,011 & 440,625 & 350,625 & 404,375 & 426,875\\
R;20;0.70;30 & 74,997 & 361,250 & 251,875 & 345,000 & 367,500\\
R;20;0.45;30 & 59,113 & 211,875 & 201,875 & 178,125 & 266,250\\
R;20;0.30;30 & 22,067 & 106,250 & 98,125 & 115,000 & 142,500\\
R;20;1.00;60 & 61,191 & 315,625 & 260,000 & 259,375 & 278,750\\
R;20;1.00;80 & 44,650 & 233,125 & 171,875 & 225,000 & 226,250\\
R;20;1.00;95 & 51,897 & 231,875 & 206,250 & 196,875 & 286,875\\
R;20;1.00;115 & 24,271 & 144,375 & 118,125 & 115,625 & 145,000\\
C;30;1.00;60 & 510,995 & 177,500 & 155,000 & 185,000 & 301,250\\
C;30;0.70;60 & 367,880 & 156,250 & 141,250 & 127,500 & 220,000\\
C;30;0.45;60 & 372,534 & 220,000 & 261,250 & 240,000 & 541,250\\
C;30;0.30;60 & 211,845 & 147,500 & 178,750 & 160,000 & 230,000\\
C;30;1.00;120 & 515,736 & 143,750 & 156,250 & 203,750 & 180,000\\
C;30;1.00;180 & 532,005 & 271,250 & 120,000 & 146,250 & 246,250\\
C;30;1.00;240 & 335,864 & 188,750 & 82,500 & 82,500 & 186,250\\
C;30;1.00;360 & 253,083 & 128,750 & 71,250 & 73,750 & 101,250\\
R;30;1.00;10 & 754,551 & 958,750 & 466,250 & 585,000 & 955,000\\
R;30;0.70;10 & 642,278 & 650,000 & 503,750 & 453,750 & 862,500\\
R;30;0.45;10 & 550,319 & 673,750 & 392,500 & 460,000 & 1,383,750\\
R;30;0.30;10 & 276,878 & 391,250 & 435,000 & 315,000 & 403,750\\
R;30;1.00;30 & 809,535 & 666,250 & 445,000 & 321,250 & 1,025,000\\
R;30;0.70;30 & 583,814 & 658,750 & 478,750 & 326,250 & 1,023,750\\
R;30;0.45;30 & 471,105 & 405,000 & 381,250 & 256,250 & 503,750\\
R;30;0.30;30 & 155,643 & 236,250 & 246,250 & 160,000 & 330,000\\
R;30;1.00;60 & 581,726 & 536,250 & 278,750 & 502,500 & 448,750\\
R;30;1.00;80 & 486,374 & 403,750 & 357,500 & 287,500 & 587,500\\
R;30;1.00;95 & 341,693 & 331,250 & 306,250 & 373,750 & 571,250\\
R;30;1.00;115 & 150,554 & 201,250 & 210,000 & 170,000 & 245,000\\
\hline
\end{tabular}
\end{table}

\begin{table}\centering
\caption{Average values of $d_{1}$. For each instance, the best
obtained value is marked with \dag.}\label{tbl_d1}
\begin{tabular}{lrrrrr} \hline
$\alpha;\beta;\gamma;\delta$ & MOLSD & PMX & OBX & UOBX & UOBX$\wedge$2EX\\
\hline
C;20;1.00;60 & \dag{ }0.0200 & 0.0426 & 0.0219 & 0.0394 & 0.0330\\
C;20;0.70;60 & \dag{ }0.0614 & 0.3471 & 0.0769 & 0.2320 & 0.0632\\
C;20;0.45;60 & \dag{ }0.0983 & 2.2945 & 1.2784 & 0.4368 & 0.3701\\
C;20;0.30;60 & \dag{ }0.1725 & 2.1288 & 0.8188 & 0.7016 & 0.7386\\
C;20;1.00;120 & \dag{ }0.0312 & 0.0494 & 0.0554 & 0.0572 & 0.0478\\
C;20;1.00;180 & \dag{ }0.0449 & 0.0510 & 0.0671 & 0.0527 & 0.0655\\
C;20;1.00;240 & \dag{ }0.0111 & 0.0793 & 0.0478 & 0.0590 & 0.0441\\
C;20;1.00;360 & \dag{ }0.5074 & 3.3814 & 2.6878 & 3.2138 & 2.6519\\
R;20;1.00;10 & 0.0950 & 0.0595 & 0.0655 & 0.0631 & \dag{ }0.0478\\
R;20;0.70;10 & 0.0768 & 0.0596 & 0.0542 & 0.0573 & \dag{ }0.0487\\
R;20;0.45;10 & 0.0701 & 0.0822 & 0.0654 & 0.0602 & \dag{ }0.0546\\
R;20;0.30;10 & 0.1396 & 0.2071 & 0.1585 & 0.1527 & \dag{ }0.1284\\
R;20;1.00;30 & 0.1025 & 0.0852 & 0.0700 & 0.0698 & \dag{ }0.0571\\
R;20;0.70;30 & 0.0841 & 0.0680 & 0.0716 & 0.0663 & \dag{ }0.0567\\
R;20;0.45;30 & \dag{ }0.1015 & 0.1502 & 0.1408 & 0.1233 & 0.1089\\
R;20;0.30;30 & 0.5280 & 0.8925 & 0.5055 & 0.4632 & \dag{ }0.4859\\
R;20;1.00;60 & 0.1137 & 0.1284 & \dag{ }0.0767 & 0.0931 & 0.0770\\
R;20;1.00;80 & 0.1238 & 0.1257 & 0.0834 & 0.0986 & \dag{ }0.0775\\
R;20;1.00;95 & 0.1029 & 0.1806 & 0.1196 & 0.1135 & \dag{ }0.0935\\
R;20;1.00;115 & 0.1913 & 0.2984 & 0.1986 & 0.2170 & \dag{ }0.1379\\
C;30;1.00;60 & \dag{ }0.0497 & 0.2773 & 0.2727 & 0.2824 & 0.1825\\
C;30;0.70;60 & \dag{ }0.3522 & 1.7077 & 1.8106 & 1.6318 & 0.4321\\
C;30;0.45;60 & \dag{ }0.1827 & 3.3573 & 1.5968 & 1.4822 & 0.8985\\
C;30;0.30;60 & \dag{ }0.2526 & 2.7933 & 1.4744 & 1.6240 & 0.9231\\
C;30;1.00;120 & \dag{ }0.1395 & 0.2371 & 0.2716 & 0.2172 & 0.1863\\
C;30;1.00;180 & \dag{ }0.1459 & 0.8585 & 0.2958 & 0.5522 & 0.4910\\
C;30;1.00;240 & \dag{ }0.1516 & 0.7634 & 0.3881 & 0.6771 & 0.2882\\
C;30;1.00;360 & \dag{ }0.1818 & 1.8217 & 1.5078 & 1.1559 & 0.9665\\
R;30;1.00;10 & 0.1031 & 0.1505 & 0.1021 & 0.1065 & \dag{ }0.0638\\
R;30;0.70;10 & 0.0991 & 0.1876 & 0.1123 & 0.1276 & \dag{ }0.0742\\
R;30;0.45;10 & 0.0999 & 0.2734 & 0.1349 & 0.1387 & \dag{ }0.0705\\
R;30;0.30;10 & \dag{ }0.1634 & 0.5062 & 0.2881 & 0.4721 & 0.1880\\
R;30;1.00;30 & 0.0826 & 0.1450 & 0.1097 & 0.0962 & \dag{ }0.0661\\
R;30;0.70;30 & 0.0858 & 0.1986 & 0.1220 & 0.0956 & \dag{ }0.0458\\
R;30;0.45;30 & \dag{ }0.1468 & 0.5630 & 0.2426 & 0.2039 & 0.1472\\
R;30;0.30;30 & \dag{ }0.2486 & 0.8227 & 0.4819 & 0.3566 & 0.3642\\
R;30;1.00;60 & 0.0960 & 0.3288 & 0.1678 & 0.0985 & \dag{ }0.0744\\
R;30;1.00;80 & \dag{ }0.1147 & 0.4261 & 0.1841 & 0.2325 & 0.1229\\
R;30;1.00;95 & \dag{ }0.0911 & 0.4424 & 0.1438 & 0.1450 & 0.0971\\
R;30;1.00;115 & \dag{ }0.1865 & 0.9801 & 0.2856 & 0.3311 & 0.2580\\
\hline
\end{tabular}
\end{table}

\begin{table}\centering
\caption{Average values of $d_2$. For each instance, the best
obtained value is marked with \dag.}\label{tbl_d2}
\begin{tabular}{lrrrrr} \hline
$\alpha;\beta;\gamma;\delta$ & MOLSD & PMX & OBX & UOBX & UOBX$\wedge$2EX\\
\hline
C;20;1.00;60 & \dag{ }0.0627 & 0.0770 & 0.0647 & 0.0857 & 0.0772\\
C;20;0.70;60 & 0.2312 & 0.5312 & 0.2436 & 0.4079 & \dag{ }0.1927\\
C;20;0.45;60 & \dag{ }0.2369 & 2.5922 & 1.5318 & 0.6826 & 0.5644\\
C;20;0.30;60 & \dag{ }0.3744 & 2.5031 & 1.1377 & 1.0032 & 1.0189\\
C;20;1.00;120 & \dag{ }0.0837 & 0.1059 & 0.1111 & 0.1184 & 0.1022\\
C;20;1.00;180 & 0.1575 & \dag{ }0.1034 & 0.1680 & 0.1662 & 0.1643\\
C;20;1.00;240 & \dag{ }0.0904 & 0.1172 & 0.1127 & 0.1306 & 0.1153\\
C;20;1.00;360 & \dag{ }0.6995 & 3.7799 & 3.1016 & 3.6282 & 3.0673\\
R;20;1.00;10 & 0.3134 & 0.1921 & 0.2469 & 0.2344 & \dag{ }0.2061\\
R;20;0.70;10 & 0.2937 & \dag{ }0.1687 & 0.2000 & 0.2190 & 0.1950\\
R;20;0.45;10 & 0.2757 & 0.2075 & \dag{ }0.2028 & 0.2252 & 0.2093\\
R;20;0.30;10 & 0.2814 & 0.4148 & 0.3113 & 0.2997 & \dag{ }0.2636\\
R;20;1.00;30 & 0.2735 & 0.1790 & 0.1792 & 0.1838 & \dag{ }0.1743\\
R;20;0.70;30 & 0.2721 & \dag{ }0.1862 & 0.2355 & 0.2535 & 0.2255\\
R;20;0.45;30 & \dag{ }0.2342 & 0.3502 & 0.2829 & 0.2589 & 0.2538\\
R;20;0.30;30 & 0.9088 & 1.3208 & 0.8674 & \dag{ }0.8331 & 0.8422\\
R;20;1.00;60 & 0.2517 & 0.2660 & 0.2001 & 0.2203 & \dag{ }0.1942\\
R;20;1.00;80 & 0.3326 & 0.2609 & 0.2530 & 0.2553 & \dag{ }0.2287\\
R;20;1.00;95 & 0.2302 & 0.3510 & 0.2472 & 0.2423 & \dag{ }0.2273\\
R;20;1.00;115 & 0.3660 & 0.5899 & 0.4476 & 0.4894 & \dag{ }0.3303\\
C;30;1.00;60 & \dag{ }0.1887 & 0.3939 & 0.3600 & 0.3769 & 0.2674\\
C;30;0.70;60 & \dag{ }0.7243 & 2.2233 & 2.3094 & 2.1416 & 0.7733\\
C;30;0.45;60 & \dag{ }0.4626 & 3.9027 & 2.0615 & 1.9792 & 1.3238\\
C;30;0.30;60 & \dag{ }0.5225 & 3.1728 & 1.8258 & 2.0087 & 1.2447\\
C;30;1.00;120 & \dag{ }0.2944 & 0.3682 & 0.4176 & 0.3649 & 0.3124\\
C;30;1.00;180 & \dag{ }0.2499 & 1.1020 & 0.4593 & 0.7660 & 0.7074\\
C;30;1.00;240 & \dag{ }0.2251 & 0.9629 & 0.5415 & 0.8705 & 0.4282\\
C;30;1.00;360 & \dag{ }0.3634 & 2.2197 & 1.9335 & 1.5391 & 1.3466\\
R;30;1.00;10 & 0.3281 & 0.4167 & 0.3338 & 0.3596 & \dag{ }0.3279\\
R;30;0.70;10 & \dag{ }0.3200 & 0.4613 & 0.3667 & 0.3417 & 0.3417\\
R;30;0.45;10 & \dag{ }0.4423 & 0.6875 & 0.5080 & 0.5000 & 0.5000\\
R;30;0.30;10 & \dag{ }0.7625 & 1.1250 & 1.0000 & 1.1299 & 1.0000\\
R;30;1.00;30 & \dag{ }0.2323 & 0.2537 & 0.2589 & 0.2663 & 0.2500\\
R;30;0.70;30 & \dag{ }0.3103 & 0.4167 & 0.3338 & 0.3333 & 0.3333\\
R;30;0.45;30 & \dag{ }0.7619 & 1.1358 & 1.0000 & 1.0000 & 1.0000\\
R;30;0.30;30 & \dag{ }0.8483 & 1.3313 & 1.0411 & 1.0000 & 1.0184\\
R;30;1.00;60 & \dag{ }0.4700 & 0.6415 & 0.5089 & 0.5000 & 0.5000\\
R;30;1.00;80 & \dag{ }0.4056 & 0.7595 & 0.5000 & 0.5280 & 0.5000\\
R;30;1.00;95 & \dag{ }0.4362 & 0.7333 & 0.4609 & 0.5000 & 0.5000\\
R;30;1.00;115 & \dag{ }0.8460 & 1.5504 & 1.0000 & 1.0285 & 1.0000\\
\hline
\end{tabular}
\end{table}

\end{appendix}

\end{document}